\documentclass[acmsmall]{acmart}
\usepackage[linesnumbered,ruled]{algorithm2e}
\usepackage{todonotes}
\newtheorem*{remark}{Fact}

\AtBeginDocument{%
  \providecommand\BibTeX{{%
    \normalfont B\kern-0.5em{\scshape i\kern-0.25em b}\kern-0.8em\TeX}}}

\setcopyright{acmcopyright}
\copyrightyear{2018}
\acmYear{2022}
\acmDOI{10.1145/1122445.1122456}

\acmJournal{TKDD}
\acmVolume{37}
\acmNumber{4}
\acmArticle{111}
\acmMonth{8}

\begin{document}

\title{Neural Network Pruning as Spectrum Preserving Process}\footnote{Preprint.}


\author{Shibo Yao}
\affiliation{%
  \institution{New Jersey Institute of Technology}
  \city{Newark}
  \state{NJ}
  \country{USA}
}
\email{espoyao@gmail.com}

\author{Dantong Yu}
\affiliation{%
  \institution{New Jersey Institute of Technology}
  \city{Newark}
  \state{NJ}
  \country{USA}
}
\email{dtyu@njit.edu}

\author{Ioannis Koutis}
\affiliation{%
  \institution{New Jersey Institute of Technology}
  \city{Newark}
  \state{NJ}
  \country{USA}
}
\email{ikoutis@njit.edu}

\begin{abstract}
Neural networks have achieved remarkable performance in various application domains.  Nevertheless, a large number of weights in pre-trained deep neural networks prohibit them from being deployed on smartphones and embedded systems.  It is highly desirable to obtain lightweight versions of neural networks for inference in edge devices.  Many cost-effective approaches were proposed to prune dense and convolutional layers that are common in deep neural networks and dominant in the parameter space.  
However, a unified theoretical foundation for the problem mostly is missing.   In this paper, we identify the close connection between matrix spectrum learning and neural network training for dense and convolutional layers and argue that weight pruning is essentially a matrix sparsification process to preserve the spectrum. Based on the analysis, we also propose a matrix sparsification algorithm tailored for neural network pruning that yields better pruning result. We carefully design and conduct experiments to support our arguments. Hence we provide a consolidated viewpoint for neural network pruning and enhance the interpretability of deep neural networks by identifying and preserving the critical neural weights. 
\end{abstract}

\ccsdesc[500]{Computing methodologies~Spectral methods}
\ccsdesc[500]{Computing methodologies~Linear algebra algorithms}
\ccsdesc[500]{Computing methodologies~Neural networks}

\keywords{neural network pruning, interpretability, matrix sparsification, randomized algorithm}

\maketitle

\section{Introduction}
Deep neural network pruning\cite{gong2014compressing} \cite{han2015learning} \cite{han2015deep} \cite{li2016pruning} has been an essential topic in recent years due to the emerging desire of efficiently deploying pre-trained models on light-weight devices, for example, smartphones, edge devices (Nvidia Jetson and Raspberry Pi), and Internet of Things (IoTs).  Neural network pruning dates back to last century when the initial attempts were made by  optimal brain damage \cite{lecun1990optimal} and optimal brain surgeon \cite{hassibi1993second}, and have achieved impressive results. 

A larger body of work, namely,  neural network compression \cite{cheng2017survey}\cite{kim2015compression}\cite{denton2014exploiting}\cite{lebedev2014speeding}, aims at removing a large number of parameters without significantly deteriorating the performance while benefiting from the reduced storage footprints for pre-trained networks and computing power.
Because the dense layers and convolutional layers usually dominate the space and time complexity in neural networks, multiple approaches have been proposed to compress these two types of network.  These approaches demonstrated surprising simplicity and superior efficacy in many situations of neural network pruning. 

Nonetheless, existing works mostly are experiment-oriented and fall into distinct categories. Thoroughly studying these related work revealed several questions.  
Can the same approaches be applied to both dense layers and convolutional layers? What are the theory and mechanisms justifying the chosen pruning operations and providing the performance guarantee of the pruned networks?  And can we gain a better insight to interpret deep learning models \cite{samek2017explainable}\cite{du2019techniques} 
from studying the topics of neural network pruning? 

In this paper, we employ spectral theory and matrix sparsification techniques to provide an alternative perspective for neural network pruning and attempt to answer the questions mentioned above. Our contributions are as follows:

\begin{itemize}
    \item We discover the relationship between neural network training and the spectrum learning process of weight matrices. By tracking the evolution of matrix spectrum, we formalize neural network pruning as a spectrum preserving process.
    \item We illustrate in detail the resemblance between dense layer and convolutional layer and essentially they both are matrix multiplication.  Consequently, a unified viewpoint, namely matrix sparsification, is proposed to prune both types of layer.  
    \item Based on our analysis, we show the potential of customizing matrix sparsification algorithm for better neural network pruning by proposing a tailored sparsification algorithm. 
    \item We thoroughly conduct experimental tasks, each of which 
    targets a specific argument to provide appropriate and solid empirical support.
\end{itemize}

The outline of this paper is as follows:  Section 2 briefly reviews related literature. In section 3, we formulate the problem of neural network pruning and explain how we tackle it from spectral theory perspective; Section 4 provides matrix sparsification techniques that are suitable for neural network pruning. 
In section 5, we generalize the analysis schema to convolutional layers.
Section 6 presents detailed empirical study; and finally the conclusions are offered. 

\section{Related Work}

The essential objective of neural network pruning is to remove parameters from pre-trained models without incurring significant performance loss. Removing parameters from the weight matrix is equivalent to eliminate neuron connections from a neural layer. Our discussion is mainly concerned with two research topics: neural network pruning and matrix sparsification.

\subsection{Neural Network Pruning}
Early attempts in pruning neural networks dated back to \textit{optimal brain damage} \cite{lecun1990optimal} and \textit{optimal brain surgeon} \cite{hassibi1993second} where the authors employed the second-order partial derivative matrix to decide the importance of connections and remove those unimportant ones. The simple yet very effective magnitude-based approach was examined in \cite{han2015learning} for both dense layers and convolutional layers, where small entries in terms of absolute value were removed from the network. The work was further expanded \cite{han2015deep} with the quantization techniques,which was previously introduced by \cite{gong2014compressing}. Channel-wise pruning \cite{li2016pruning} for convolution was also examined based on the $l_1$-norm magnitude of the filters. 
The lottery ticket hypothesis \cite{frankle2018lottery} and a consequent work \cite{zhou2019deconstructing}
have again attracted the attention of the community on the magnitude-based pruning.  


There is another line of efforts \cite{denton2014exploiting} \cite{vasilescu2002multilinear} \cite{lebedev2014speeding} \cite{kim2015compression} \cite{liu2015sparse} based on low-rank approximation, which has a similar taste to pruning but with a more clear theoretical support, falling within a larger scope of work namely neural network compression \cite{cheng2017survey}.  The low rank approximation was applied to the weight matrix or tensors \cite{kolda2009tensor} in order to reduce the storage requirements or inference time for large pre-trained models. Some other works on neural network compression also include weight sharing \cite{chen2015compressing} and hash trick \cite{chen2016compressing}, where they also look at the problem in the frequency domain.



\subsection{Matrix Sparsification}


Matrix sparsification is important in many numerical problems, e.g. low-rank approximation, semi-definite programming and matrix completion, which widely exist in data mining and machine learning problems. Matrix sparsification is to reduce the number of nonzero entries in a matrix without altering its spectrum. The original problem is NP-hard \cite{mccormick1983combinatorial}\cite{gottlieb2010matrix}. The study of approximation solutions to this problem was pioneered by \cite{achlioptas2007fast}, and further expanded in \cite{achlioptas2013near} \cite{arora2006fast} \cite{achlioptas2013matrix} \cite{nguyen2009matrix} \cite{drineas2011note}. An extensive study on the error bound was done in \cite{gittens2009error}. 

Since the spectrum of the sparsified matrix does not deviate significantly from that of the original matrix, serving as a linear operator the matrix retains its functionality,  i.e. 
$A\in \mathbb{R}^{m\times n}$ is a mapping $\mathbb{R}^m \rightarrow \mathbb{R}^n$. 
We can define the matrix sparsification process as the following  optimization problem: 
\begin{equation}
\begin{aligned}
& \underset{}{\text{min}}
& & \| \Tilde{A} \|_{0} \\
& \text{s.t.}
& & \| A - \Tilde{A}\| \leq \epsilon \\
\end{aligned}
\label{eqn:linear-combine}
\end{equation}
where $A$ is the original matrix, $\Tilde{A}$ is the sparsified matrix, 
$\|\cdot\|_{0}$ is the $0$-norm that equals the number of non-zero entries in a matrix, $\|\cdot\|$ denotes matrix norm, $\epsilon \geq 0$ is the error tolerance. 

In matrix sparsification, we often use the spectral norm (2-norm) $\|\cdot\|_{2}$ and the Frobenius norm (F-norm) $\|\cdot\|_{F}$ to measure the deviation of the sparsified matrix from the original one. 



\section{Problem Formulation}
Given a dense layer $\textbf{z}_{t} = \sigma (\textbf{z}_{t-1}^T A + \textbf{b})$,  where $\textbf{z}_{t-1} \in \mathbb{R}^m$ is the input signal, 
$\textbf{z}_{t} \in \mathbb{R}^n$ is the output signal, 
$A \in \mathbb{R}^{m\times n}$ is the weight matrix, 
$\textbf{b} \in \mathbb{R}^n$ is the bias, $\sigma$ denotes some activation function, we desire to obtain a sparse version of $A$ denoted by $\Tilde{A}$ such that $A$ and $\Tilde{A}$ have similar spectral structure and $\textbf{z}_{t-1}^T \Tilde{A}$ is as close to $\textbf{z}_{t-1}^T A$ as possible. 


Similarly for a convolution $z_t = T * z_{t-1}$ we want to find a sparse version of $T$ such that the convolution result is as close as possible, where $T$ could be a vector, matrix or higher-order array depending on the order of input signal and the number of output channel. By closeness, we use norms as metric. We start from the investigation on dense layers and then make generalization on convolutional layers.

\subsection{Neural Net Training as Spectrum Learning Process}
In a dense layer, we focus on the $\textbf{z}^T A$ part since it contains most parameters.
Neural network is essentially a function simulator that learns some artificial features,
which is achieved by linear mappings, nonlinear activations, and some other customized units
(e.g. recurrent unit).
For the linear mapping, the analysis is usually done on the spectral domain. 

Recall that Singular Value Decomposition (SVD) is optimal under both spectral norm \cite{eckart1936approximation} and Frobenius norm \cite{mirsky1960symmetric}. 
The weight matrix $A \in \mathbb{R}^{m\times n}$ as a linear operator can be decomposed as
$$
A = U\Sigma V^T = \sum_{i=1}^{min(m,n)}\sigma_i \textbf{u}_i \textbf{v}_i^T
$$
where $U = [\textbf{u}_1,\textbf{u}_2,...,\textbf{u}_m] \in \mathbb{R}^{m \times m}$ is 
the left singular matrix, 
$V = [\textbf{v}_1, \textbf{v}_2,..., \textbf{v}_n] \in \mathbb{R}^{n\times n}$ is the right singular matrix and and $\Sigma$ contains the singular values $diag(\sigma_1, \sigma_2,..., \sigma_n)$
in a non-increasing order.

Note that a input signal $\textbf{z} \in \mathbb{R}^m$ can be written into
a linear combination of $\textbf{u}_i$, i.e. $\textbf{z}_i = \sum_i^m c_i \textbf{u}_i$ where
$c_i$ are the coefficients.
Thus, the mapping from $\textbf{z} \in \mathbb{R}^m$ to $\textbf{z}^{\prime} \in \mathbb{R}^n$ is
$$
\textbf{z}^{\prime} = \textbf{z}^TA 
= \sum_{i=1}^m c_i\textbf{u}_i^T \sum_{j=1}^{min(m,n)}\sigma_j \textbf{u}_j \textbf{v}_j^T
= \sum_{j=1}^{min(m,n)} c_j\sigma_j \textbf{v}_j
$$
where $\sigma_j$ are non-increasingly ordered, since
$\textbf{u}_i^T\textbf{u}_j=0, \forall i\neq j$
and $\textbf{u}_i^T\textbf{u}_i=1, \forall i$.


We are especially interested in finding out how the spectrum $\Sigma$, i.e., singular values, of the linear mapping evolve during neural network training. As a matter of fact,  our empirical study shows that the neural network training process is a spectrum learning process. To investigate the spectral structure, we design the following task:


\textbf{Task 1}: check the spectra and norms of dense layer weight matrices and how they change during neural network training.

\begin{figure*}[ht]
  \centering
  \includegraphics[width=.8\textwidth]{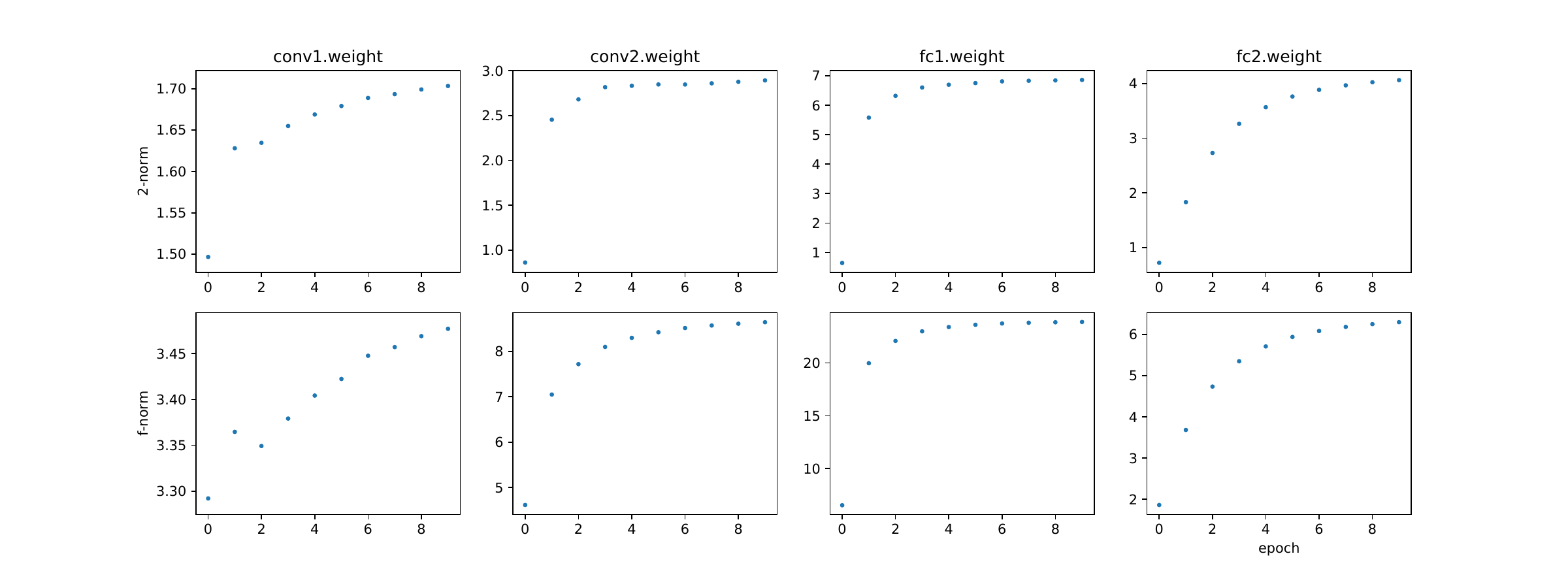}
  \caption{(Task1) Matrix Norms in LeNet stabilize during training.
  y-axis denotes matrix norm and x-axis denotes training epoch.}
  \label{norm learning 1}
\end{figure*}

\begin{figure*}[ht]
  \centering
  \includegraphics[width=.9\textwidth]{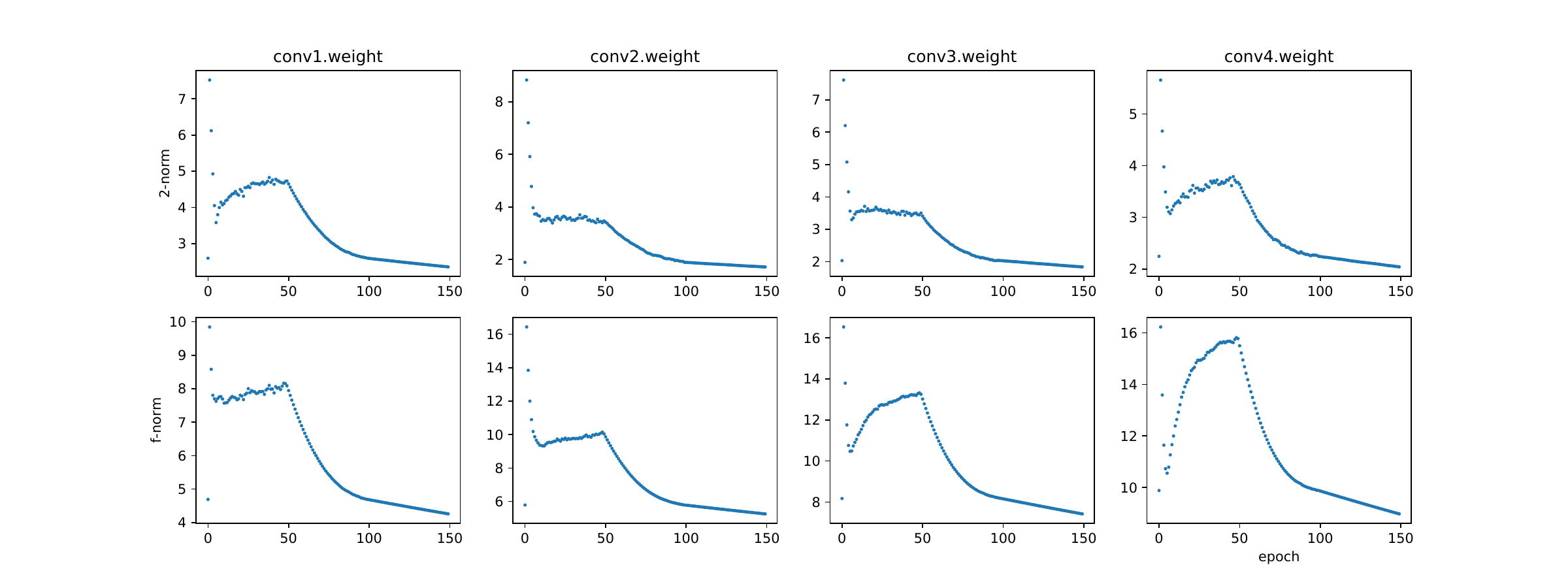}
  \caption{(Task1) Matrix Norms in VGG19 stabilize during training.}
  \label{norm learning 2}
\end{figure*}

\subsection{Neural Net Pruning as Spectrum Preserving Process}


Following the logic, if the neural network training is a spectrum learning process, when we practise neural network pruning, we would like to preserve the spectrum in order to preserve the neural network performance. In other words, we want to obtain a sparse $\Tilde{A}$ that has similar singular values to $A$. How to measure the wellness of spectrum preservation? We can use the spectral norm (2-norm) $\|A\|_2 = \sigma_1$ which is the largest singular value since we care about the dominant principle component, and the Frobenius norm (F-norm)
$$
\|A\|_F = (\sum_{i=1}^m\sum_{j=1}^n A_{ij}^2)^{1/2} = (Tr(A^TA))^{1/2} = (\sum_{i=1}^{min(m,n)} \sigma_i^2)^{1/2}
$$
which is usually considered an aggregation of the whole spectrum. 
Note that $\|A\|_2 \leq \|A\|_F$ and $\|A\|_F \leq \sqrt{min(m,n)}\|A\|_2$. 

Therefore the goal is to find a sparse $\Tilde{A}$ such that 
$\|A - \Tilde{A}\|_2 \leq \epsilon$ or $\|A - \Tilde{A}\|_F \leq \epsilon$. 
To show that the pruning process is a spectrum preserving process, we design the following task. 

\textbf{Task2}: apply magnitude-based thresholding at different sparsity levels and check the relationship between the resulting weight matrix spectra and the performance of the pruned neural networks.

\begin{figure}[ht]
  \centering
  \includegraphics[width=.6\textwidth]{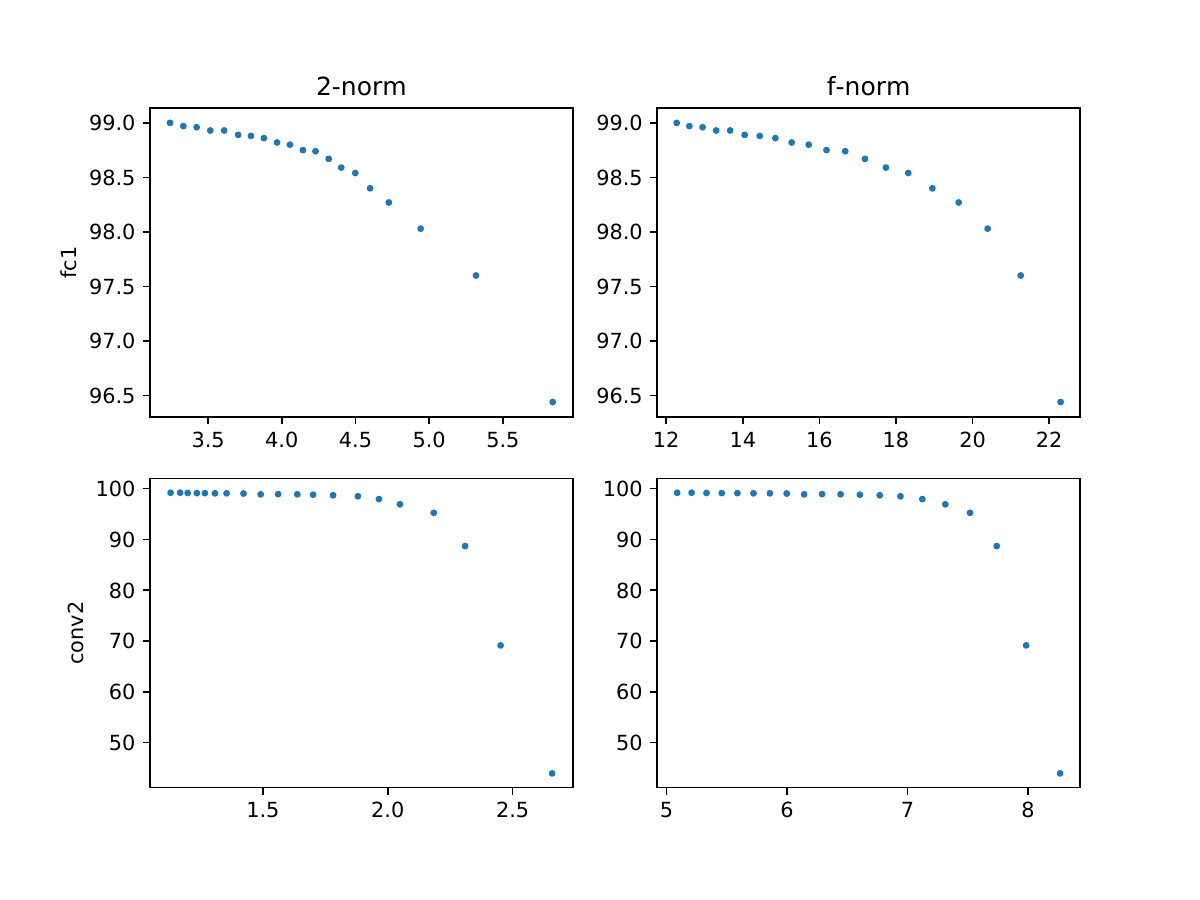}
  \caption{(Task2) $\|A-\Tilde{A}\|_2$ and $\|A-\Tilde{A}\|_F$ V.S. 
  neural network accuracy as the sparsity increases (LeNet on MNIST). 
  As we increase sparsity, the $2$-norm  $\|A-\Tilde{A}\|_2$ and the $F$-norm $\|A-\Tilde{A}\|_F$ increase while  the neural network performance drops accordingly. 
  y-axis denotes accuracy and x-axis denotes matrix norm.}
  \label{lenet sparsity acc}
\end{figure}

\begin{figure}[ht]
  \centering
  \includegraphics[width=.6\textwidth]{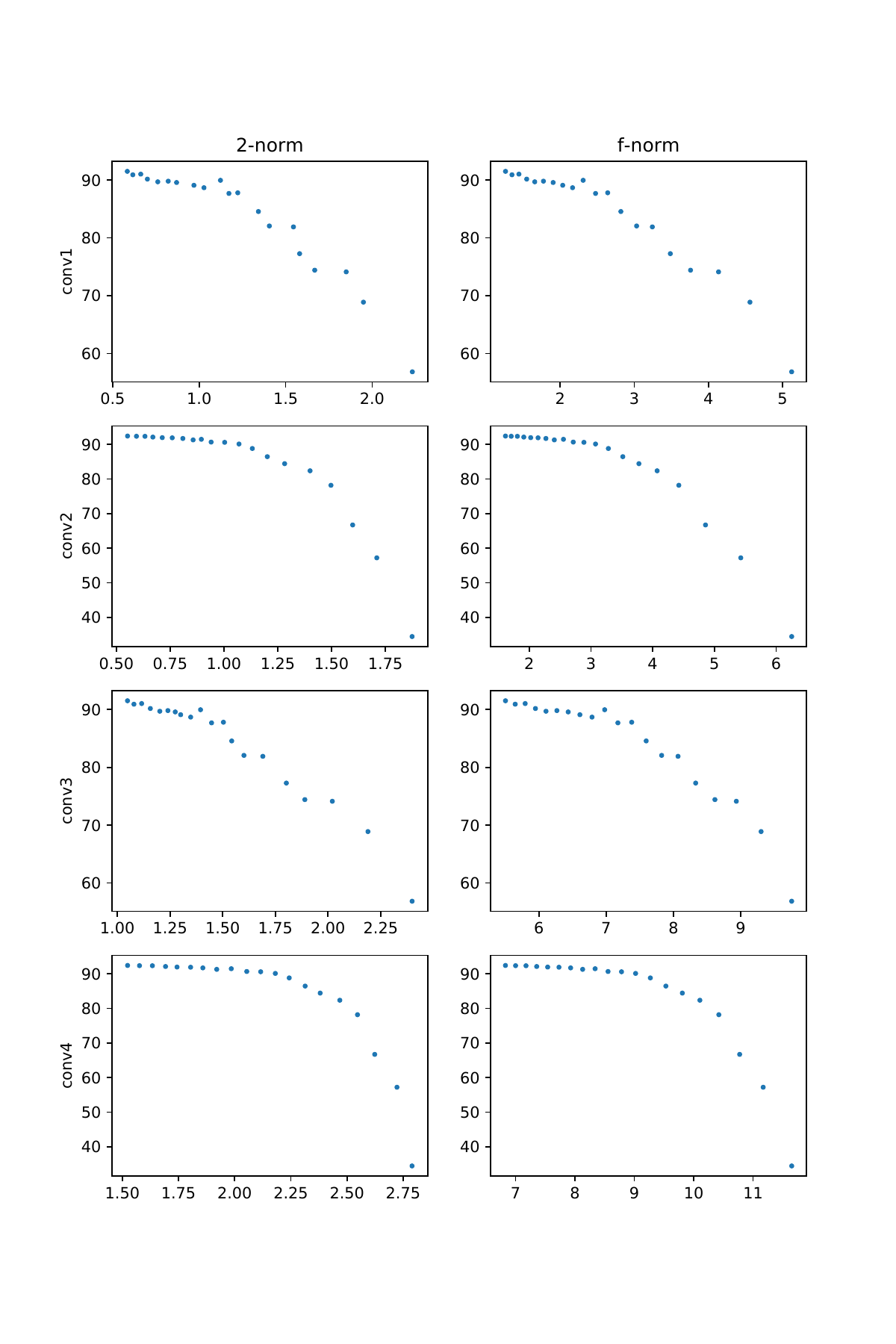}
  \caption{(Task2) $\|A-\Tilde{A}\|_2$ and $\|A-\Tilde{A}\|_F$ V.S. 
  neural network accuracy as sparsity increases (VGG19 on CIFAR10).
  The axis labels are the same as those in Figure \ref{lenet sparsity acc}.}
  \label{vgg sparsity acc}
\end{figure}

\subsection{Iterative Pruning and Retraining}
From the optimization perspective, the pretrained neural network achieved satisfying local optimum. Once the weight matrix is sparsified, the spectrum to some extent deviates and the optimality no longer holds. We wish to retrain the neural network such that it returns to the satisfying local optimum and the performance is preserved. Apparently we don't want the neural net deviate from the local optimum too far, otherwise it would be difficult to return to the optimum due to the nonconvex optimization process of neural network. Hence, iterative pruning and retraining has been a common practice for neural network pruning. We also design a task to elaborate from the spectrum perspective. 

\textbf{Task}3: Inspect the weight matrix spectra in each pruning iteration, 
i.e. the weight matrix spectra after pruning and after retraining respectively, and compare the corresponding neural network performances. We
also include a comparison to one-shot pruning. 

\begin{figure*}[ht]
  \centering
  \includegraphics[width=\textwidth]{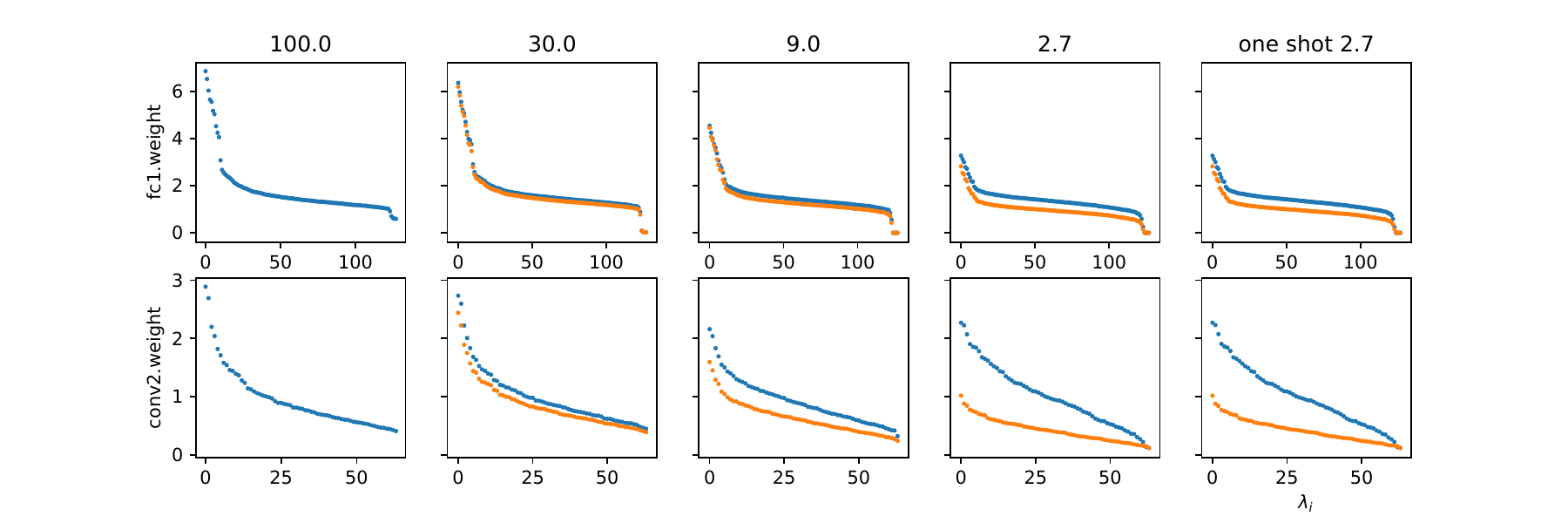}
  \caption{(Task3) Iterative pruning for LeNet on MNIST. 
  The number above each column denotes the percentage of parameter left in a weight matrix. 
  The right-most are the one-shot pruning results.
  We can see that the spectra deviate after pruning (orange) and tend to recover after retraining (blue).
  Once the sparsity reaches a certain point, the spectrum collapses 
  which corresponds to model performance significant drop. }
  \label{iter pruning}
\end{figure*}

\begin{figure*}[ht]
  \centering
  \includegraphics[width=\textwidth]{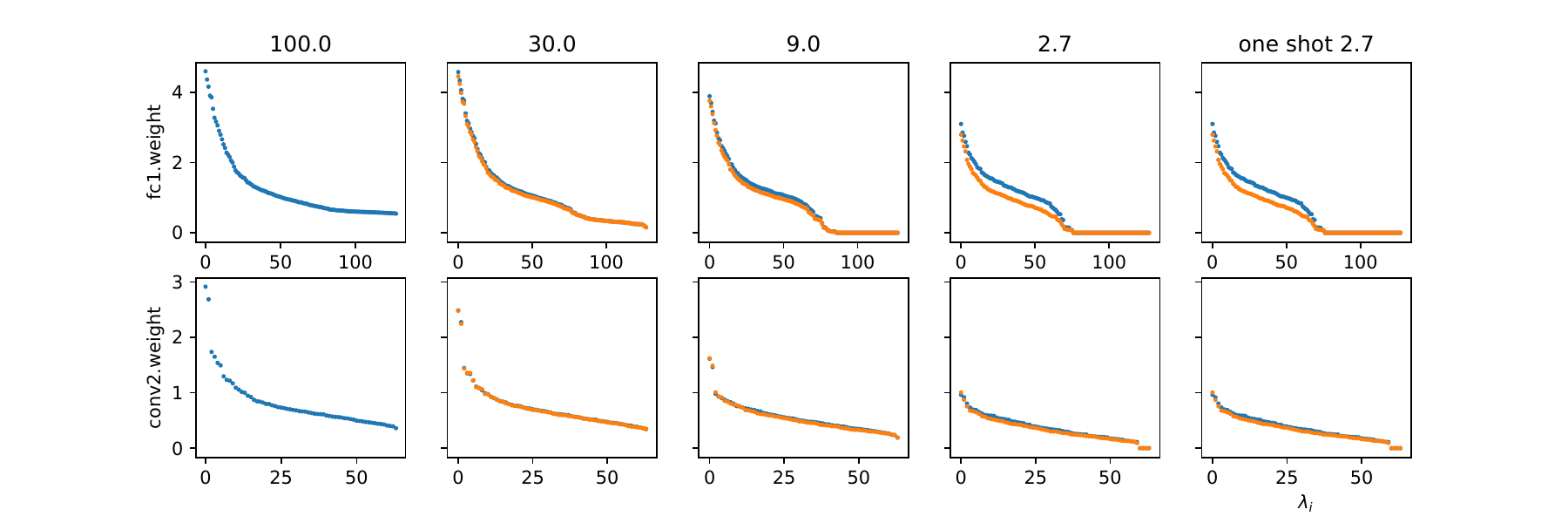}
  \caption{(Task3) Iterative pruning for LeNet with Batch Normalization on MNIST. 
  Comparing to LeNet without Batch Normalization, the spectra recovery behaviour is insignificant. }
  \label{iter pruning3}
\end{figure*}

\section{Matrix Sparsification Algorithms}

In the previous section, we identify the relationship between neural network pruning (removing parameters/connections) and the spectrum preserving process. Matrix sparsification plays a primary role in the spectrum preserving process and will guide the network pruning process. In this section, we present the practical techniques of matrix sparsification. 

\subsection{Magnitude Based Thresholding}

Magnitude-based neural network pruning have attracted a lot of attention and show supprisingly simplicity and superior efficacy. 
In the context of matrix sparsification, this is a straightforward approach, namely magnitude-based matrix sparsification or hard thresholding. 
Given a matrix $A$, let $\Tilde{A}$ denote its sparsifier. Entry-wise we have
\[\Tilde{A}_{ij} = \begin{cases}
                A_{ij} &  |A_{ij}| > t \\
                0 &   else
                \end{cases}
\].
\begin{remark}\label{f_opt}
Magnitude based thresholding always achieves sparsification optimality in terms of F-norm. 
\end{remark}
The fact can be trivially verified since using $|A_{ij}|$ and using $A_{ij}^2$
(on which F-norm is based)
are equivalent in terms of deciding small entries in a matrix.
However throwing away small entries does not always guarantee
the optimal sparsification result in terms of 2-norm. And in many situations,
we care more about the dominant singular value instead of the whole spectrum.


\subsection{Randomized Algorithms}
In randomized matrix sparsification, each entry is sampled according to some distribution independently and then rescaled. E.g. each entry is sampled according to a Berboulli distribution, and we either set it to zero or rescale it. 
\[\Tilde{A}_{ij} = \begin{cases}
                A_{ij}/p_{ij} &  p_{ij} \\
                0 &   1-p_{ij}
                \end{cases}
\]
where $p_{ij}$ can be a constant or positively correlated to the magnitude of the entry.
The following theorem provides the justification to this type of matrix sparsification.

\begin{theorem}\label{the1}
A matrix where each entry is sampled from a zero-mean bounded-variance distribution possesses weak spectrum with large probability. 
\end{theorem} 
By weak spectrum, it means small matrix norm. To be more concrete, since matrix norm is a metric and triangle inequality applies,
we have
$$
\|A\| \leq \|\Tilde{A}\| + \|A-\Tilde{A}\|
$$
. We need to show that
$N = A-\Tilde{A}$ falls within the category of matrices described in Theorem \ref{the1}. Since
$$
E(N_{ij}) = E(A_{ij} - \Tilde{A}_{ij}) = A_{ij} - A_{ij}/p_{ij}\cdot p_{ij} = 0
$$
and
$$
var(N_{ij}) = var(\Tilde{A}_{ij}) = (A_{ij}/p_{ij})^2 \cdot p_{ij} = A_{ij}^2 / p_{ij}
$$
, as long as $A_{ij}^2 / p_{ij}$ is upper-bounded, which is true most of time,
$var(N_{ij})$ is bounded. Therefore the
randomized matrix sparsification can guarantee the error bound.

\subsection{Customize Matrix Sparsification Algorithm for Neural Network Pruning}
In this section, we propose a customized matrix sparsification algorithm to show the potential of 
designing a better spectrum preservation process in neural network pruning. We do not intend to
present a new state-of-the-art neural network pruning algorithm. There are two important points in our proposed algorithm: truncation and sampling based on the principal components of  explicit truncated SVD.
To the best of our knowledge,   sampling based on probability proportional to principal components is employed for the first time in designing matrix sparsification for neural network pruning . 

First, we adopt the truncation trick that is common in existing work. As clearly pointed out by \cite{achlioptas2013matrix}, the spectrum of the random matrix $N = A - \Tilde{A}$ is determined by its variance bound. Usually, the larger the variance, the stronger the spectrum of the random matrix. Existing works took advantage of the finding and proposed truncation \cite{arora2006fast}\cite{drineas2011note} in sparsification, i.e. to set small entries to zero while leaving large entries as is and sampling on the remaining ones. 
\[\Tilde{A}_{ij} = \begin{cases}
                A_{ij} & |A_{ij}| > t \\
                0  & p_{ij} < c \\
                A_{ij}/p_{ij}\cdot Bern(p_{ij}) & else 
                \end{cases}
\]
where $p_{ij} \propto |A_{ij}|$, $t$ is decided by the quantile (leave large entries as is), 
and $c$, the lower threshold for zeroing weights, as a constant could be set manually, 
and $Bern(\cdot)$ denotes Bernoulli distribution.

Second, instead of sampling based on the probability calculated from the magnitude of the original matrix entry, 
we do sampling based on the probability calculated from the principal component matrix entry magnitude
with a little compromise on complexity, 
in order to better preserve the dominant singular values. 
Matrix sparsification was originally proposed for fast low-rank approximation on very large matrices, 
due to the fact that sparsity accelerates matrix-vector multiplication in power iteration. Essentially, 
we desire to find the sparse sketch $\Tilde{A}$ of $A$ that preserves the dominant singular values well. 
This coincides with the goal of layer-wise neural network pruning from the spectrum preserving viewpoint --
we desire to preserve the dominant singular values, based on the fact that we often consider
information lies in the low-frequency domain while noises are in the high-frequency domain. 
The major difference is that weight matrices in neural network, either from dense layers or convolutional layers,
are usually not too large, 
and therefore explicit SVD or truncated SVD on them is fairly affordable. Once we have access to the principle components
of the weight matrices, we are able to preserve them better in the sparsification process. 
Note that preserving dominant singular values is a harmonic approach between preserving the 2-norm and the F-norm, 
since $\|A\|_2 = \lim_{p\to\infty} (\sum_i \sigma_i^p)^{1/p}$ and $\|A\|_F = \lim_{p\to2} (\sum_i \sigma_i^p)^{1/p}$.

The crutial part is to find the low-rank approximation $B$ to $A$, where 
$
B = \sum_{i=1}^K \sigma_i\textbf{u}_i\textbf{v}_i^T
$ and $\sigma_i \textbf{u}_i \textbf{v}_i^T$ are from SVD on A.
We set the entry-wise sampling probability based on $|B_{ij}|$, i.e. $p_{ij} \propto |B_{ij}|$. 
Alg \ref{Sparsify} presents the sparification algorithm. The \textit{partition} function is the one used in \textit{quicksort}. 
\begin{algorithm}[]
    \SetKwFunction{Union}{Union}\SetKwFunction{FindCompress}{FindCompress}\SetKwInOut{Input}{input}\SetKwInOut{Output}{output}
    
    \Input{($A$, $c$, $q$, $K$)} 
    \Output{$\Tilde{A}$} 
    \BlankLine  \tcp*{$q$: quantile above which remain unchanged}
    $B =$ \textit{truncated-SVD($A$, $K$)}; \tcp*{$B$: low-rank approx to $A$}
    $m,n$ = shape of $B$; \\
    $t$ = partition($\{|B_{ij}|\}$, int($m\times n \times q$)); \\
    \For{$i\gets1$ \KwTo $m$}{
        \For{$j\gets1$ \KwTo $n$}{
            \uIf{$|B_{ij}| < t$}{
                $p_{ij} = (B_{ij}/t)^2$; \\
                \uIf{$p_{ij} < c$}{
                    $A_{ij} = 0$; \\
                }
                \uElse{
                    $A_{ij} = A_{ij}/p_{ij}\cdot Bern(p_{ij})$; \\
                }
            }
        }
    }
    \caption{Sparsify}\label{Sparsify}
\end{algorithm}

\begin{figure*}[ht]
  \centering
  \includegraphics[width=\textwidth]{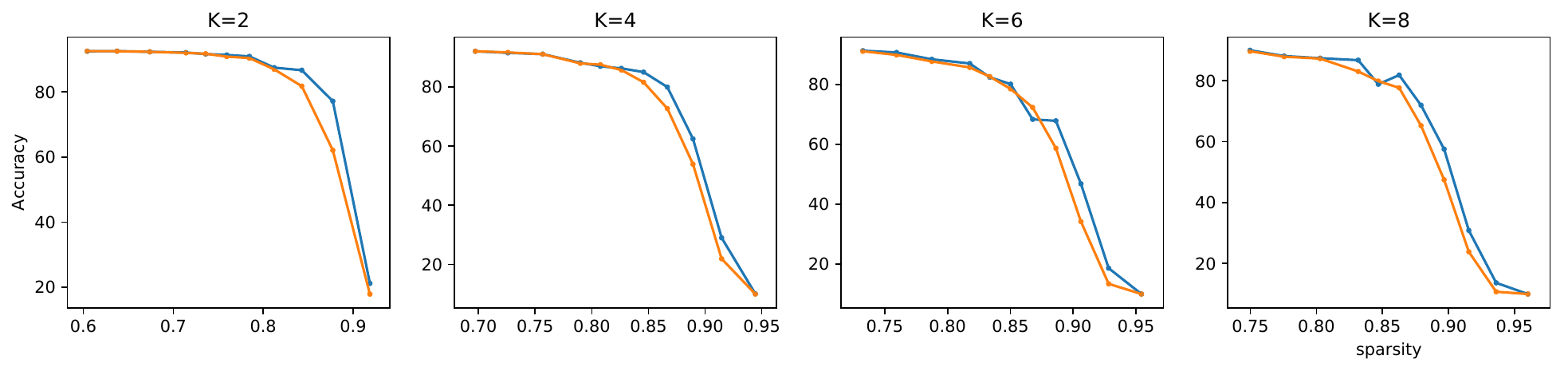}
  \caption{(Task4) Pruned Network Testing Performance given by Magnitude-based Thresholding (orange) v.s. Algorithm\ref{Sparsify} (blue). 
  The larger $q$, the larger sparsity.}
  \label{acc compare}
\end{figure*}

We need to demonstrate that the proposed sparsification algorithm preserves dominant singular values better and improves the generalization performance of the pruned network.  We propose the following task to check whether it makes improvement based on our analysis.

\textbf{Task 4} Apply the above algorithm on VGG19 layer-wise weight matrix sparsification, compare the generalization
performance of the pruned network and the pruned network given by thresholding at the same sparsity level. 

Here we provide a high level proof on sparsification error being upper bounded. 
Let $D$ denote the sparse sketch generated by setting smallest entries in $A$ to 0 and $\Tilde{A}$ as usual the final sparsfied result. 
From Fact \ref{f_opt} we know
that $D$ is the optimal sketch of $A$ in terms of F-norm, i.e. $\|A-D\|_F = \epsilon_*$.
Based on Theorem \ref{the1} and its illustration we know that 
$N=\Tilde{A}-D$ satisfies the zero-mean
and bounded-variance condition. Hence $\|\Tilde{A}-D\|_F \leq \epsilon_0$. Therefore 
if we apply triangle equality given matrix norm is a metric, 
$$
\|A-\Tilde{A}\|_F \leq \|A-D\|_F + \|D-\Tilde{A}\|_F \leq \epsilon_* + \epsilon_0. 
$$
Some other techniques, e.g. quantization\cite{gong2014compressing}\cite{han2015deep}, 
can be used together  with 
sparsification to further compress matrices and neural networks. Essentially they are 
also spectrum preservation techniques \cite{achlioptas2007fast}\cite{arora2006fast}.

\section{Generalization to Convolution}

\begin{figure*}[ht]
  \centering
  \includegraphics[width=0.9\textwidth]{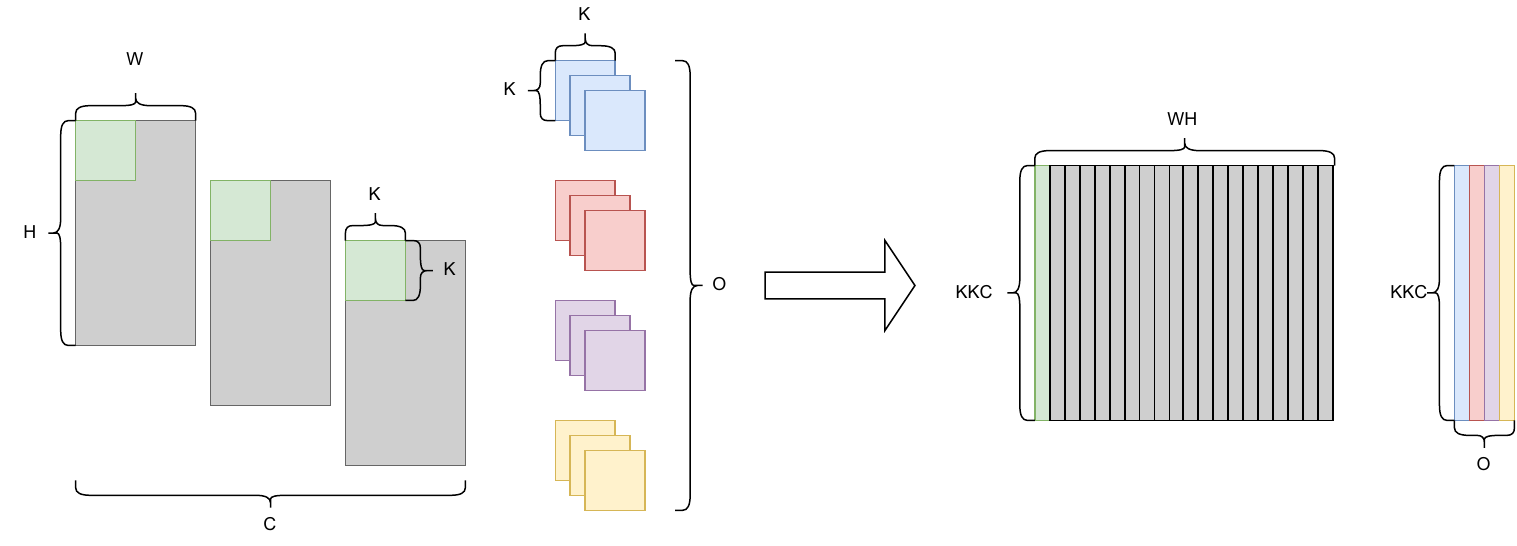}
  \caption{Convolution as Dense Matrix Multiplication}
  \label{cnn as mat}
\end{figure*}

Extensive literatures argue that convolutional layers compression can be formalized as tensor algebra problems \cite{lebedev2014speeding}\cite{denton2014exploiting} \cite{kim2015compression}\cite{liu2015sparse}\cite{sun2016sparsifying}. 
However, it's advantageous to explain convolutional layer pruning from the matrix viewpoint since the linear algebra have many nice properties that do not hold for multilinear algebra. We want to ask: can we still provide theoretical support to convolutional layer pruning 
using linear algebra we have discussed so far?

\subsection{Pruning on Convolutional Filters}
In this section we state and illustrate the following fact. 
\begin{remark}\label{CNN mat mult}
Discrete convolution in neural networks can be represented by dot product between two dense matrices. 
\end{remark}

To see this, suppose we have a convolutional layer with input signal size of $W\times H$ as width by height, and with $C$ input channels and $O$ output channels. Here we consider a 2-d convolution on the signal. The kernel is of size $C \times K \times K$ and there are $O$ such kernels. For the sake of simplicity in notations, suppose the striding step is 1, half-padding is applied and there is no dilation (for even $W$ and $H$ the above setting results in output signal of size $W\times H$ as width by height). 2-d convolution means that the kernel is moving in two directions. Fact \ref{CNN mat mult} has been utilized to optimize lower-level implementation of CNN on hardware \cite{chellapilla2006high}\cite{chetlur2014cudnn}. Here we take advantage of the idea to unify neural network pruning on dense layers and convolutional layers with matrix sparsification. 



Let us focus on one single output channel, one step of the convolution operation is the summation of element-wise product of two higher-order array, i.e. the kernel $G \in \mathbb{R}^{C\times K\times K}$ and 
the receptive field of the signal of the same size $X \in \mathbb{R}^{C\times K\times K}$. Note that taking the summation of element-wise product is equivalent to
vector inner product. Therefore if we unfold the kernel for a single output channel to a vector and rearrange the receptive field of the signal accordingly to another vector, a single convolution step can be treated as two vector inner product, i.e. $G*X = \textbf{g}^T\textbf{x}$ where $\textbf{g}, \textbf{x} \in \mathbb{R}^{CKK}$. Since we have $O$ output channels in total, there are $O$ such kernels of the same size. All of them being unfolded, we then can convert the convolution into a matrix product $Z^TA$, where $A \in \mathbb{R}^{CKK \times O}$ being the kernels and $Z \in \mathbb{R}^{CKK \times WH}$ being the rearranged input signals. And consequently the output signal $Y \in \mathbb{R}^{WH \times O}$ 
(as mentioned before, stride 1, half padding and no dilation result in 
input signal and output signal being in the same shape). 
Figure \ref{cnn as mat} visualizes convolution as matrix multiplication.

The matrix multiplication representation of convolution discussed above generalizes to any other convolution settings. 
Also note that the way we unfold the filters does not affect the spectrum of the resulting matrix, 
since row and column permutations do not change matrix spectrum. 
Therefore, all the analyses based on simple linear algebra we have discussed so far generalize to convolutional layer pruning. 
To verify our analyses,

\textbf{Task1,2,3} will also be conducted on convolutional layers. 
When we say weight matrix in a context of convolution, 
it refers to as the matrix unfolded from the convolution higher-order array in the way described in Figure \ref{cnn as mat}.

\subsection{Convolutional Filter Channel Pruning}

\begin{figure}[ht]
  \centering
  \includegraphics[width=0.8\textwidth]{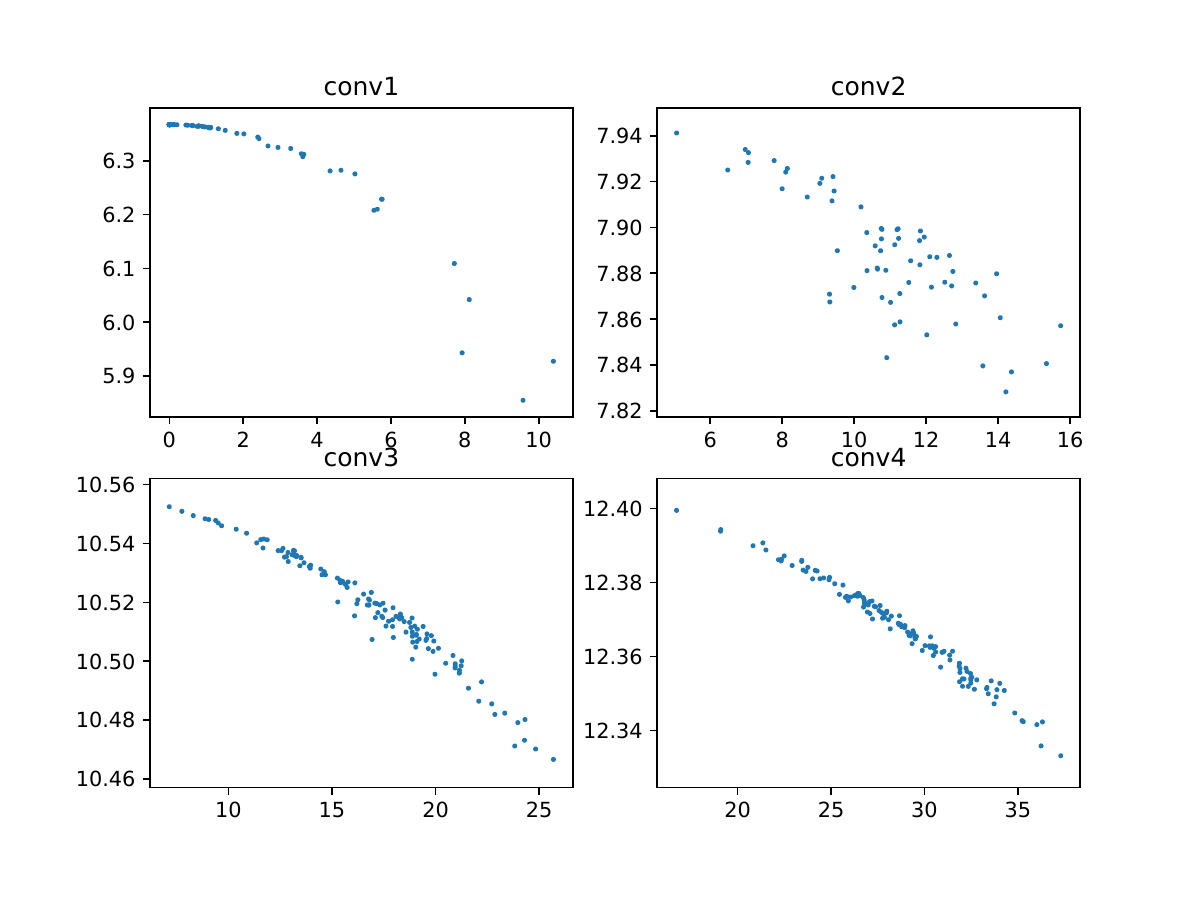}
  \caption{(Task5) VGG19 Channel Pruning based on $\sum_{i}|T_i|$. 
  y-axis denotes $\|\Tilde{A}\|_F$ and x-axis denotes $\sum_{i}|T_i|$. 
  The smaller $\sum_{i}|T_i|$, the smaller $\|A-\Tilde{A}\|_F$, 
  the larger $\|\Tilde{A}\|_F$, the better spectra are preserved.
  Hence our analysis bridges the gap between small $\sum_{i}|T_i|$ 
  and good neural network performance preservation.}
  \label{channel pruning}
\end{figure}


Entry-wise pruning almost always results in unstructured sparsity
that requires specific data structure design in network deployment in order
to realize the complexity reduction from pruning. Therefore it's desirable 
to prune entire channels from convolutional layers to achieve higher efficiency.  There is another important work on pruning channels \cite{li2016pruning}. The approach is to take small $\sum_{ijk}|T_{ijk}|$ where $T$ denotes the filter for a specific channel. This is equivalent to remove a column in $A$ we just discussed with small-magnitude values. It's also a spectrum preserving process as $\sum_{ijk}|T_{ijk}|$ is fairly a proximity to $\sum_{ijk}(T_{ijk})^2$ on which the F-norm is based. Hence pruning the whole filter with small $\sum_{ijk}|T_{ijk}|$ is to preserve the F-norm of the convolution matrix we discussed in the previous subsection. To check the relation between $\sum_{ijk}|T_{ijk}|$ and the F-norm of $\Tilde{A}$, we propose the following task. 

\textbf{Task5} Pruning different filters and check the relationship between $\sum_{ijk}|T_{ijk}|$ 
and the resulted convolution matrix F-norm $\|\Tilde{A}\|_F$.

\section{Empirical Study Details}
In this section, we present the proposed task details and results. 
The experiments are mainly based on LeNet \cite{lecun1998gradient} on MNIST and VGG19 \cite{simonyan2014very} on CIFAR10 dataset \cite{krizhevsky2009learning}.  We trained the neural networks from scratch based on the official PyTorch \cite{paszke2019pytorch} implementation.  Then we conducted our experiments based on the pre-trained neural networks. 

We trained LeNet with a reduced number of epochs of 10. All other hyperparameter settings are the ones used in the original implementation of the PyTorch example. The VGG19 was trained with the following hyperparameter setting: batch size 128, momentum 0.9, weight decay $5e^{-4}$,  and the learning rates of 0.1 for 50 epochs, 0.01 of 50 epochs, and 0.001 of another 50 epochs.  The final testing accuracy for LeNet on MNIST and VGG19 on FICAR10 was 99.14\% and 92.66\%, respectively. We saved the layer weight matrices for each epoch of training, including two fully connected layers and two convolutional layers for LeNet and the first four convolutional layers for VGG19 (due to the limitation in the reporting space). We then study the evolvement of 2-norm and F-norm of weight matrices during training. 

In Figures \ref{norm learning 1} and  \ref{norm learning 2}, we observe that the 2-norm and F-norm of a particular weight matrix change fast at the beginning of training and tend to stabilize as training proceeds.  This observation provides concrete evidence that network training is essentially a spectrum learning process. Note that the initial spectrum is not necessarily flat (see Figures \ref{spectrum learning 1} and \ref{spectrum learning 2} in Appendix), but rather depends on the initialization. The stabilization also has its explanation from the optimization perspective: as training goes on we are trapped into a satisfying local optimum and the gradients are almost zero 
for layers when chain rule applied, which means the weight matrices are not being updated significantly.  

\textbf{Task 2} We investigated the relationship between spectrum preservation and the performance of the pruned neural network.  Matrix sparsification (hard thresholding) was employed to prune neural networks.  We varied the percentage of parameter preservation from 20\% to 1\% to get different sparsities (the sparser,  the larger $\|A-\Tilde{A}\|_2$ and $\|A-\Tilde{A}\|_F$). We pruned different layers in the pre-trained LeNet and VGG19 and checked their performance without retraining. 

Figure \ref{lenet sparsity acc} and figure \ref{vgg sparsity acc} show that when $\|A-\Tilde{A}\|_2$ increases, the neural network performance deteriorates  almost monotonically.  It is also true for $\|A-\Tilde{A}\|_F$.  The finding confirms that the better the spectrum of weight matrix is
preserved during pruning, the better the performance of the pruned neural network performance is conserved. It is also interesting to note that the experiment observations (data points) tend to concentrate on the upper left side, rather than the lower right of plots, and there is a drop-off in each plot.  The pattern indicates that the spectrum and neural network performance collapse when the sparsity of the matrix sketch goes beyond a certain point.

\textbf{Task 3} We also inspected the spectra of the weight matrices during the iterative pruning and retraining.  Specifically, we adopted the magnitude-based pruning (hard thresholding matrix sparsification) to retain 30\% parameters in each iteration. We conducted experiments on the second convolutional layer and the first dense layer which contain most parameters in LeNet.  Once the parameters were pruned, we applied a mask on that certain matrix
to fix the zero values which correspond to the pruned connections between neurons. We also applied masks on all other
layers during retraining in order to get a better assess on certain weight matrix spectrum change due to sparsification. 
The retraining process was done with two additional epochs. 

Figure \ref{iter pruning} shows a clear pattern of how the spectrum of a weight matrix evolves once we iteratively prune it. The first plot in each row is the spectrum of the original weight matrix. The orange dots denote the spectra of the weight matrices after each round of pruning, and the blues dots represent the spectra after retraining in each iteration. 
In the beginning, we can prune a large number of parameters from the weight matrix without altering the spectrum significantly, and we still can recover the spectrum to some extent by retraining. 
When the sparsity reaches a certain point, the spectrum seems to collapse and fail to recover its original shape and position even with retraining, which is consistent with the general observation neural network performance significantly drops when it's too sparse.

Such a spectrum recovery behaviour is negligibly insignificant on VGG19. The spectra deviate from the original ones in pruning yet are rather slightly modified during retraining. This is mainly due to the widely adopted Batch Normalization\cite{ioffe2015batch}, 
which rescales training batches to zero-mean unit-variance
batches and hence significantly eliminates the need for shaping the matrix spectra.  In the beginning, we managed to remove the batch normalization that comes after each convolutional layer
in VGG19. However, this leads to a difficult training situation for deep neural networks, as the community is widely aware of \cite{ioffe2015batch}. Hence, we adopted an alternative approach to demonstrate batch normalization
effect on the learning process of spectra. We added batch normalization after each convolutional layer in LeNet, repeated the experiment aforementioned, and compared the results. For a detailed treatment of batch normalization, please refer to\cite{ioffe2015batch}\cite{bjorck2018understanding}.

In Figure \ref{iter pruning3}, we observe that when batch normalization is applied,  the spectrum recovery is less significant than that of without batch normalization, although both the trajectory and the ending status of the spectra for iterative pruning and one-shot pruning
surprisingly reassemble to each other.

\textbf{Task 4} We applied algorithm\ref{Sparsify} on all convolutional layers in VGG19 at the same time, varied algorithm
settings to get different sparsities, recorded the corresponding testing performance of the pruned network, 
and compared with the performance of the pruned network via thresholding at the same sparsity level. 
Due to the randomness in our proposed algorithm, the sparsity in different layers is also different. 
We present the aggregated sparsity, i.e. the total number of nonzero parameters divided by total number of parameters 
in all convolutional weight matrices, in our empirical study result. To ease the implementation and focus on 
our arguments, we fixed parameter $c=0.5$, varied the quantile parameter $q$ and the number of principal components $K$. 

From Figure \ref{acc compare} we can see that, our proposed algorithm almost always leads to better pruned network 
generalization performance without retraining compared to that given by thresholding, at different sparsity levels.
This demonstrates the potential of designing and customizing matrix sparsification algorithms 
for better neural network pruning approaches. In addition, we also observed Alg \ref{Sparsify} almost always 
yield smaller sparsification error compared to thresholding in terms of 2-norm, which is exactly the motivation of the algorithm design.

\textbf{Task 5} To support the channel pruning mechanism interpretation, we inspected the relation 
between $\sum_i|T_i|$ and $\|\Tilde{A}\|_F$. For this task, we checked the first four convolutional 
layers in VGG19. Based on our illustration on convolutional layer as dense matrix multiplication, 
we unfolded each channel filter from a 3rd-order array to a vector, and then concatenated all the channel
vectors into a matrix. Therefore removing a filter channel is equivalent to removing a column in 
the so called convolution matrix in figure \ref{cnn as mat}. We have shown the relation between spectrum 
preservation and neural network performance preservation. As long as we can show the relation between 
$\sum_i|T_i|$ and $\|\Tilde{A}\|_F$, we can bridge the gap and associate $\sum_i|T_i|$ and neural network 
performance. 

Figure \ref{channel pruning} shows that more or less $\|\Tilde{A}\|_F$ of a convolution matrix 
is negatively correlated to $\sum_i|T_i|$, which is consistent to our analysis.

\section{Conclusion and Future Work}
In this work, we argue that neural network training has a strong interdependence with spectrum learning.  The relationship provides neural network pruning with a theorical formulation based on spectrum preserving, to be more specific, matrix sparsification.  We reviewed the existing primary efforts on neural network pruning and proposed a unified viewpoint for both dense layer and convolutional layer pruning. We also designed and conducted experiments to support the arguments, and hence provided more interpretability to deep learning related topic. 

We anticipate that the superior algorithm design for neural network pruning rests upon the effective and efficient matrix sparsification using spectral theory and lower-level implementation of sparse neural network for the targeted systems.  For future works, we will undertake further investigation on how activation functions affect pruning and more pruning algorithms on other types of network layers.

\bibliographystyle{ACM-Reference-Format}
\bibliography{citation}

\newpage\clearpage
\appendix
\onecolumn
\section{Supplementary Materials for Reproducibility}




\subsection{python code for Alg\ref{Sparsify}}
\begin{verbatim}
def lowRankSampling(A, tao=0.3, n_comp=5):
    A_ = np.array([i.flatten() for i in A]) 
    U,sigma,V = randomized_svd(A_, n_comp)
    B = np.zeros(A_.shape)
    for i in range(n_comp):
        B += np.outer(U[:,i],V[i])*sigma[i]

    flat_B = abs(B.flatten())
    thresh_B = int(flat_B.size*tao)-1
    t_B = np.partition(flat_B, thresh_B)[thresh_B]

    with np.nditer(A, op_flags=["readwrite"]) as it: 
        for x in it: 
            if abs(x) < t_B:
                p = np.square(x/t_B)
    
                if p < 0.5:
                    x[...] = 0 
                else:
                    x[...] = np.random.binomial(1,p,1) * x / p 

    return A

\end{verbatim}

\subsection{Additional Empirical Study Results}
\begin{figure*}[h]
  \centering
  \includegraphics[width=\textwidth]{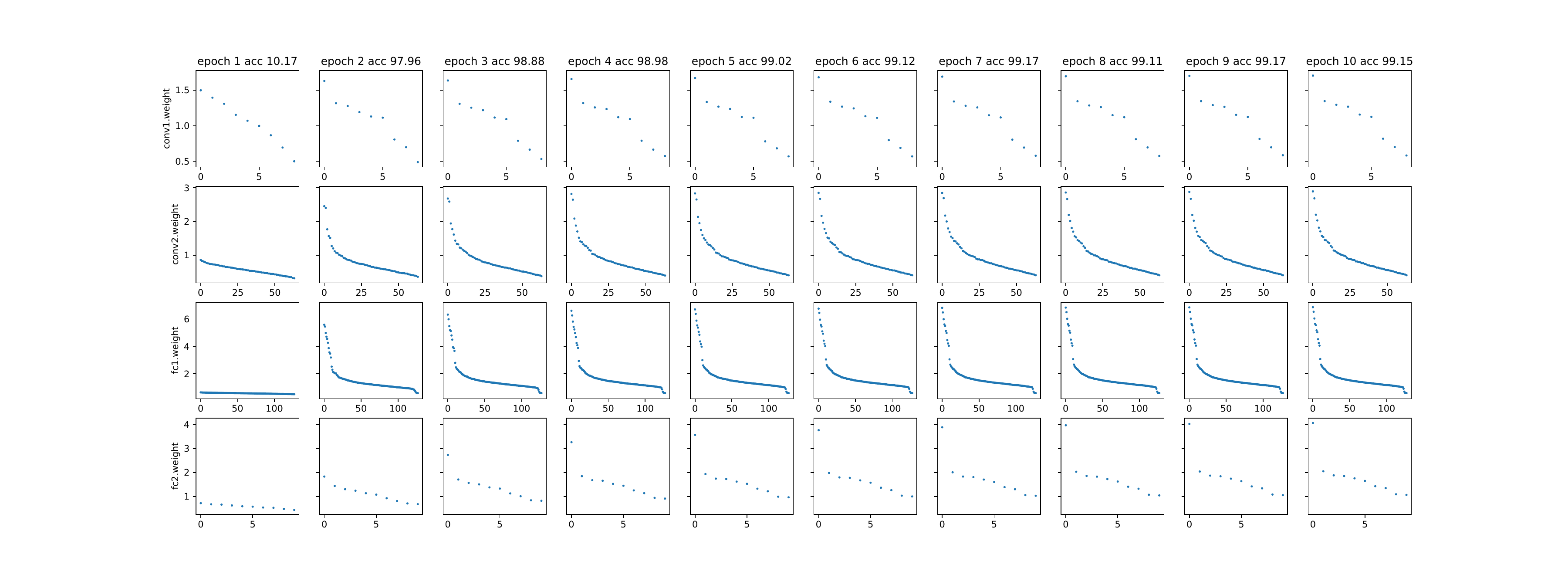}
  \caption{(Task1) Matrix Spectra in LeNet stabilize during training. 
  Each dot denotes a singular value of the weight matrix.}
  \label{spectrum learning 1}
\end{figure*}

\begin{figure*}[h]
  \centering
  \includegraphics[width=\textwidth]{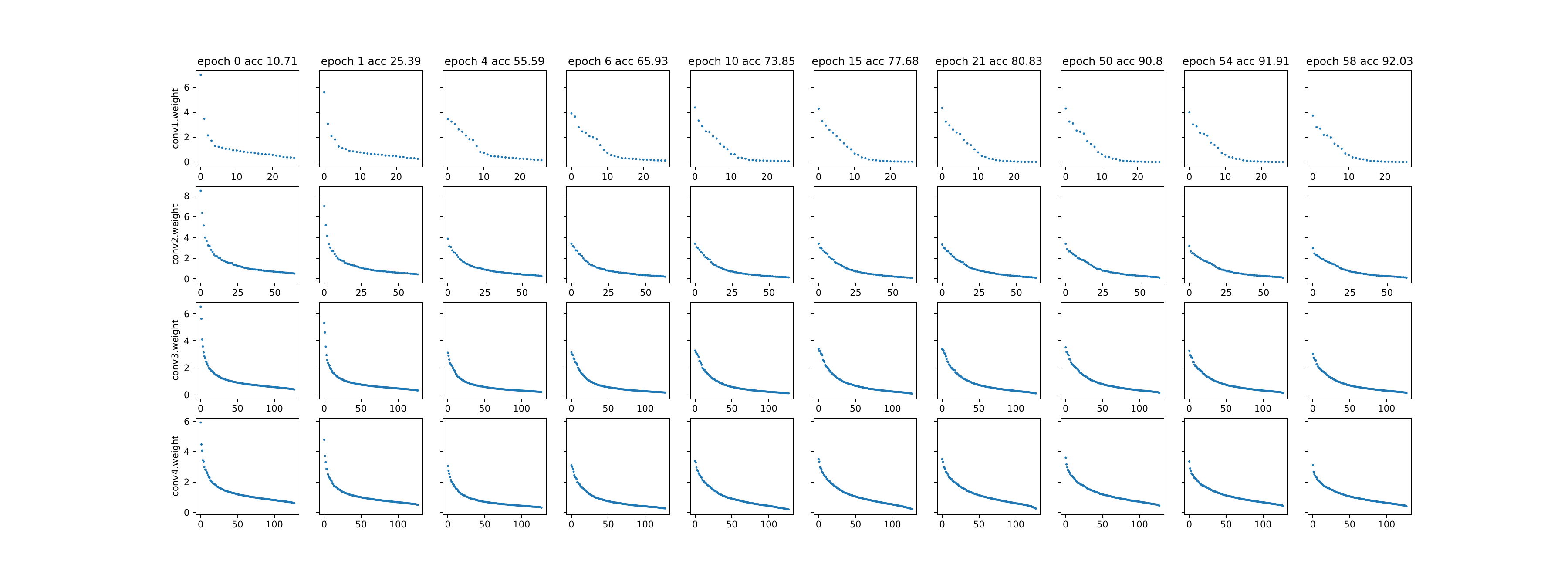}
  \caption{(Task1) Matrix Spectra in VGG19 stabilize during training.}
  \label{spectrum learning 2}
\end{figure*}

\end{document}